\documentclass[10pt, a4paper]{article}

\usepackage[final]{lrec2026} 
\usepackage{booktabs}
\usepackage{amsmath}
\usepackage{transparent}
\usepackage{array, makecell}

\usepackage[dvipsnames]{xcolor}
\title{MUSCAT: MUltilingual, SCientific ConversATion Benchmark
}

\name{
  \begin{tabular}{c}
    Supriti Sinhamahapatra$^{1}$, Thai-Binh Nguyen$^{1}$, Yiğit Oğuz$^{1}$, Enes Ugan$^{1}$, \\
     Jan Niehues$^{1}$, Alexander Waibel$^{1,2}$
  \end{tabular}
}

\address{$^{1}$Karlsruhe Institute of Technology , $^{2}$Carnegie Mellon University\\
         \{supriti.sinhamahapatra, thai-binh.nguyen, enes.ugan, jan.niehues\}@kit.edu\\ 
         yigit.oguz@student.kit.edu, alexander.waibel@cmu.edu \\}

\abstract{The goal of multilingual speech technology is to facilitate seamless communication between individuals speaking different languages, creating the experience as though everyone were a multilingual speaker. To create this experience, speech technology needs to address several challenges: Handling mixed multilingual input, specific vocabulary, and code-switching. However, there is currently no dataset benchmarking this situation. We propose a new benchmark to evaluate current Automatic Speech Recognition (ASR) systems, whether they are able to handle these challenges. The benchmark consists of bilingual discussions on scientific papers between multiple speakers, each conversing in a different language. We provide a standard evaluation framework, beyond Word Error Rate (WER) enabling consistent comparison of ASR performance across languages. Experimental results demonstrate that the proposed dataset is still an open challenge for state-of-the-art ASR systems. The dataset is available at \url{https://huggingface.co/datasets/goodpiku/muscat-eval}.
\vspace{0.3cm} 
\noindent }

\begin{document}

\maketitleabstract

\section{Introduction}
Seamless communication across language boundaries is a long-term dream of mankind. The ultimate goal is to have a natural, multilingual conversation where each participant talks in their favorite language and is able to understand all the other languages.
While significant progress has been made in terms of multilingual speech recognition in high resource \cite{barrault2023seamless, liu2023kit}, as well as low resource settings \cite{robinson2025jhu, li2025kit}, to the best of our knowledge, currently there is no realistic benchmark to evaluate systems in  multilingual dialogue scenario. 

To address the growing need for realistic and high-quality multilingual datasets, we present an unique collection of audio recordings designed as a benchmark for automatic speech recognition. 
In order to build strong systems for this benchmark, several challenges need to be addressed: multilingual speech input, speaker segmentation, audio condition, domain-specific vocabulary, and code-switching.

To collect the data, we setup a discussion between two bi-lingual speakers about scientific publications. Each speaker only speaks one language but understands both languages. 
While English is the dominant language for
scientific communication globally, we aim for AI-based solutions that can facilitate scenarios where
researchers can communicate scientific content
in their native language without any compromise.
 To study this, we created a controlled
simulation using bilingual speakers.

\begin{figure}[ht]
  \centering
  \includegraphics[width=1.0\linewidth]{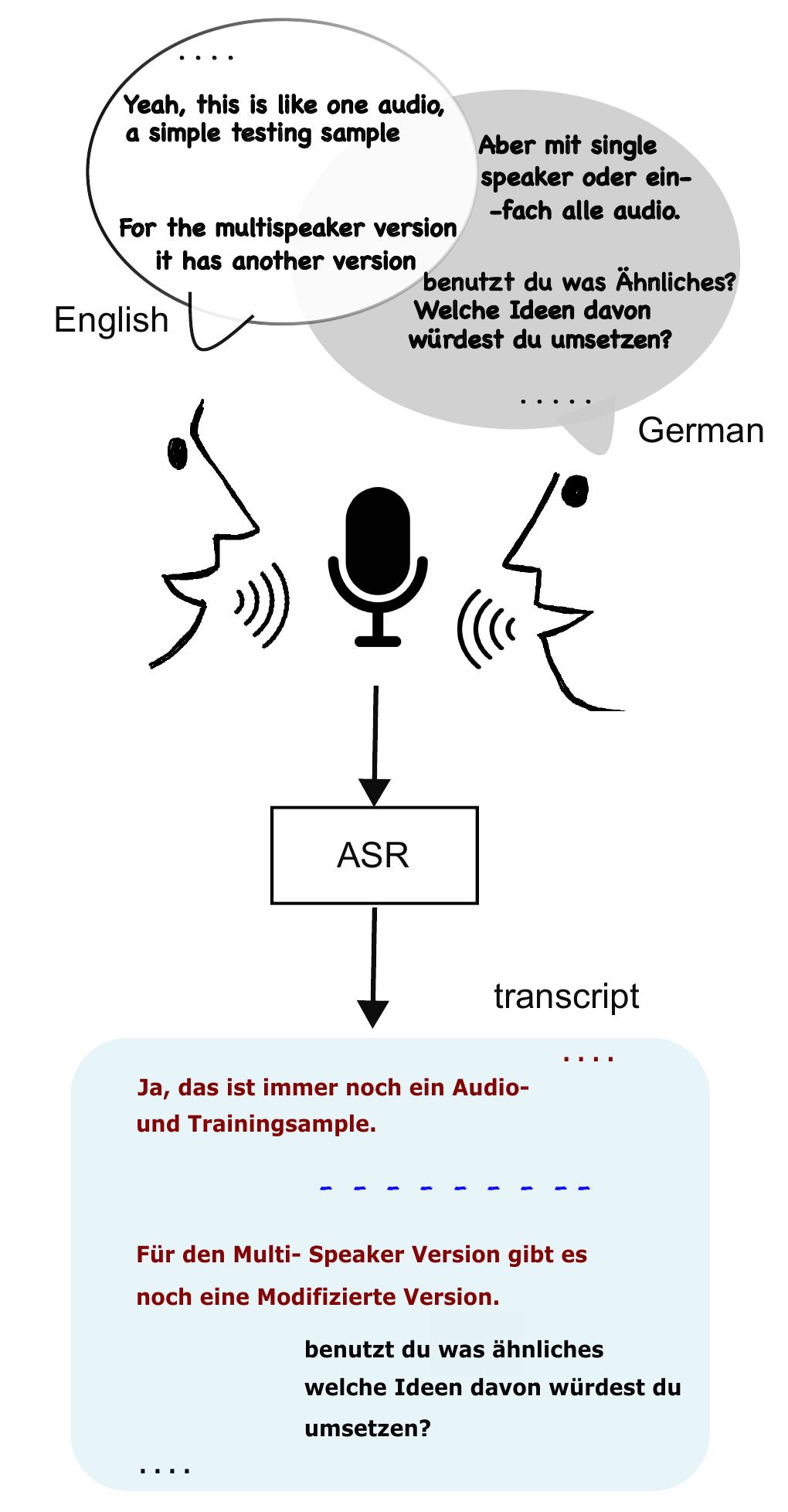}
  \caption{An example illustrating the creation of MUSCAT (upper part of the figure) and the challenges its multilingual diversity poses for state-of-the-art ASR systems (lower part of the figure). The ASR is unable to accurately detect the language switches in a spontaneous conversation denoted by red in the transcript. The blue dashed lines ($---$) represent the part of the conversation that ASR fails to transcribe.}
  \label{fig:speech_production}
\end{figure}

Our evaluation show limitation in current
AI systems. 
The technology is not yet robust to
support multilingual communication in scientific
domains.
In Figure~\ref{fig:speech_production}, the upper part illustrates an example scenario of the MUSCAT dataset creation process, in which spontaneous conversations are recorded between two speakers, one speaking English and the other German. The lower part highlights a key challenge faced by state-of-the-art (SOTA) ASR models: their difficulty in accurately detecting language switches in spontaneous multilingual speech. In such cases, the model often either translates the utterance into the language of the preceding context or fails to transcribe certain segments altogether. 
This leads to an oracle setup for speech translation technology.

The main contributions of this paper are:
\begin{itemize}
    \item A new benchmark
    \footnote{\url{https://huggingface.co/datasets/goodpiku/muscat-eval}}
    for multilingual conversations.
    \item A detailed analysis of the challenges of the proposed benchmark.
    \item State-of-the-art baseline results that highlight the difficulties of the task.
\end{itemize}
\section{Data Collection}

We aim to build a high-quality multilingual  dataset. In order to achieve this, we first create a conversation setup where the challenges of multilingual, scientific conversations are highlighted. Next, we design a recording setup that allows us to investigate the different challenges individually.


\subsection{Conversation Setup}
Each instance in the dataset contains a conversation between a pair of speakers over a scientific paper.  The speaker possess prior knowledge of the paper being discussed.
We create the oracle situation for speech
translation in this scenario by having the speakers
converse in two different languages. 
To carry out the conversation, the speakers need to be fluent in both  languages. For our case, the speakers engaging in natural conversations are each fluent in English at a C1 level and are native speakers of the other language.

This unique setup allows for meaningful exchanges where speakers fully comprehend one another but respond solely in 
one of the two languages.
These conversations offer paired speech\footnote{All participants gave their consent for their voices to be recorded and used for research purposes} and transcripts for language pairs like English-German, English-Vietnamese, English-Chinese, and English-Turkish. 



%

\subsection{Recording Setup} \label{sec:experimental_setup}
The challenges associated with ASR vary depending on the recording environment.
In order to evaluate different conditions jointly, we synchronously record the conversations with three different devices.
The first device is Meeting Owl 3, which is a popular video conferencing system that captures 360° video and audio.
The second device, the ReSpeaker USB Microphone, is a compact array microphone designed for high-fidelity, multi-directional audio capture.
The third, Aria smart glasses by Meta, function as a wearable device that records first-person audio along with audio from the speaker’s environment.

The Meeting Owl 3, referred to as simply OWL in the rest of the paper, was connected to a laptop via USB, and recordings were made using OBS Studio 3. The ReSpeaker USB Microphone Array was paired with a Raspberry Pi 3, also connected via USB, to provide an additional audio source. For brevity, we refer to this setup as Pi for the remaining of the paper.
Finally, the Aria glasses by Meta were worn by one of the speakers and used to record the entire conversation from their perspective. 
Since Aria can be worn by only one person during recording, we randomly selected one speaker. This results in three German, one Chinese, one Vietnamese, and one English speaker wearing Aria. The other audio recording devices were kept approximately at the middle and equidistant from the speakers. 

The recordings are made at a sampling rate of 44.1kHz using multiple microphones as mentioned above. After data collection, the recordings from all devices were manually aligned using Audacity to ensure they were perfectly synchronized. This combination of devices and meticulous alignment ensures that the dataset captures a wide range of audio perspectives, adding depth and variety. All software used were the latest versions available at the time of recording. To ensure minimum possible interference from external sound, an appropriately secluded room is used for the recordings.

\section{Human Annotation}

We annotate the collected data to be used as a benchmark for state-of-the-art ASR systems. In a first step, we perform a manual segmentation of the audio recordings which serves as the oracle to evaluate and compare two automatic segmentation approaches. Next, we create the multilingual transcripts of the audio.

\subsection{Manual Segmentation}\label{sec:manual_seg}
We perform the manual segmentation guided by two constraints.
First, since each of our recordings consists of conversations in two different languages, we prioritize language-specific segmentation. This ensures that each segment comprises recordings in a single language.  
Second, to support optimal model performance, we limit each segment to a maximum duration of 30 seconds. The manual segmentation process is conducted using Label Studio \cite{studio2023label}, an open-source data annotation tool.
 


\subsection{Human post-editing}
 Starting from the derived manual segments, we follow a two-step process to obtain the human transcript of our dataset. 
Firstly, to ease manual transcription of the recordings, we use a state-of-the-art ASR model, 
Whisper~\cite{radford2023robust} for automatic transcription of the language specific segments. 
As a second step, the respective speaker  manually corrects any mistakes made by ASR model, ensuring high quality transcription.
This strategy was adopted to overcome the challenge of finding external annotators who possessed both fluency in the specific languages and familiarity with the complex scientific discourse discussed in the papers. Using the speakers for such task ensured that technical terms and domain-specific context were annotated accurately. 

 During the annotation process, we frequently observe instances of code-switching within the recorded speech. Since mixed-language utterances are known to pose particular challenges for current speech recognition systems~\cite{klejch2021cstr,hamed2022investigations,ugan2025weight}, annotators were additionally instructed to mark all words belonging to the embedded language whenever code-switching occurred.

\section{MUSCAT Dataset}
The MUSCAT dataset consists of multilingual conversations of six recordings across eleven speakers. Each recording is between a pair of speakers, and there exists one speaker who is present in two recordings. All six recordings have at least one English speaker, while the other speaks one of the languages from German, Turkish, Chinese, and Vietnamese. Of the six recordings, half of the conversations are between a pair of English and German speakers, while the other half is between English and the remaining languages.  
We maintain gender diversity among the speakers in the dataset. To this end, among the eleven speakers of the MUSCAT dataset, six speakers are male and the remaining  five speakers are female. 

In order to evaluate different challenges related to this benchmark, we provided 6 different variations of the dataset. First, three different recordings with the different devices per speaker. Secondly, for each conversation, segmented and  unsegmented version of the audio recording is available.

Table~\ref{tab:dataset_statistics} summarizes the main aspects of the dataset. The total duration of our dataset is approximately 65 minutes, of which English-Vietnamese conversation comprises of 17 minutes, whereas English-Chinese, English-Turkish, and English-German conversations comprise 15, 12, and 21 minutes, respectively. Words spoken by each speaker are attributed to the word count of their respective language, which in total is 9,066 words. 
\begin{table}[ht]
  \caption{Dataset Statistics}
  \label{tab:dataset_statistics}
  \centering
  \resizebox{1.0\linewidth}{!}{%
  \begin{tabular}{l|ccc}
  
    \toprule
   
     \textbf{Recordings}      & \textbf{Languages} & \textbf{Total }      & \textbf{Total }       \\
     \textbf{}      & \textbf{} & \textbf{Duration}      & \textbf{ Word Counts}       \\
  \midrule
    Recording 1   & English    & 4.69 mins & 463  \\
                    & German     & 1.92 mins & 288  \\
      Recording 2   & English    & 1.39 mins & 162  \\
                    & German     & 2.74 mins & 427  \\
      Recording 3   & English    & 7.51 mins & 1344  \\
                    & Turkish    & 3.94 mins & 447  \\
      Recording 4   & English    & 11.90 mins & 1362  \\
                    & Chinese    & 2.79 mins & 623  \\
      Recording 5   & English    & 7.47 mins & 972  \\
                    & German     & 3.00 mins & 426  \\
      Recording 6   & English    & 10.04 mins & 1489  \\
                    & Vietnamese & 6.83 mins & 1063  \\
                     \midrule
        Total           &             &   64.22 mins  &9,066\\
    \bottomrule
  \end{tabular}}
  
\end{table}

\section{Baseline}
This section outlines the baseline configuration adopted in our experiments, detailing the ASR models used and the segmentation strategies applied during pre-processing.
\subsection{ASR Models}
Our goal is to evaluate the performance of SOTA ASR models on the MUSCAT dataset. To this end, we employ four SOTA models, Whisper, SALMONN,Phi-4 Multimodal and Wav2Vec2, and asses thequality of their generated transcriptions.These models represent diverse ASR paradigms, including encoder–decoder architectures (Whisper), multimodal large language models (SALMONN and Phi-4 Multimodal), and CTC-based systems (Wav2Vec2).

\paragraph*{Whisper}
Whisper is a transformer-based encoder-decoder model developed by OpenAI, primarily designed for automatic speech recognition (ASR) and speech translation tasks \cite{radford2023robust}. It has been trained on approximately 680k hours of speech data collected from the internet. The model's encoder processes the input speech to generate audio features, which are then passed to the decoder. The decoder, using these audio features along with positional encodings, produces the corresponding transcription. Whisper also incorporates a set of context tokens that guide the model by specifying the language, the task to be performed, and the start and end points of the transcription.
\paragraph*{SALMONN}
The SALMONN model, developed by Tsinghua University and ByteDance \cite{tang2023salmonn}, extends the capabilities of Large Language Models (LLMs), such as Vicuna \cite{chiang2023vicuna}, to directly perceive and interpret general audio inputs. This enhancement allows LLMs to perform competitively across a range of speech and audio processing tasks. SALMONN integrates information from two specialized encoders, Whisper \cite{radford2023robust} for speech and BEATs \cite{chen2022beats} for general audio using a window-level Q-Former module \cite{zhang2024vision}. The resulting augmented audio tokens are aligned with the LLM’s internal representations, enabling seamless multi-modal understanding.

\paragraph*{Phi-4-multimodal}
Phi-4-multimodal (referred to as Phi) is a 5.6B-parameter instruction-tuned multi-modal transformer developed by Microsoft. It is designed for unified processing of text, image, and audio inputs, enabling it to handle tasks across vision-language, vision-speech, and speech-language domains. The model supports a context length of up to 128K tokens and utilizes 32 transformer layers equipped with Grouped Query Attention (GQA) \cite{ainslie2023gqa} for efficient long-context processing. Vision and audio modalities are mapped into the text embedding space via two-layer multilayer perceptrons (MLPs). Phi demonstrates strong performance on a wide range of multilingual and multi-modal benchmarks.

\paragraph*{wav2vec2}
wav2vec2 \cite{baevski2020wav2vec} is a self-supervised learning framework designed to learn speech representations directly from raw audio. 
The model includes a convolutional feature encoder that converts audio into latent representations, followed by a transformer network that captures context over time.



We use the wav2vec2-large-960h-lv60-self model from Facebook, which is trained on 960 hours of audio for performing experiments with the english audios. For other languages, including German, Turkish, Chinese  and Vietnamese  we employ the wav2vec2-large-xlsr-53 model fine-tuned in a supervised manner on the respective Common Voice datasets \cite{ardila2019common}. 
The model is trained separately for each language using a Connectionist Temporal Classification (CTC) loss \cite{baevski2020wav2vec, graves2006connectionist} to perform ASR \cite{grosman2021xlsr53-large-german, wav2vec2-large-xlsr-53-turkish, grosman2021xlsr53-large-chinese, wav2vec2-large-xlsr-53-vietnamese}.


\subsection{Segmentation}
\label{subsec:segmentation}
For the condition using unsegmented audio, segmentation is necessary because some of the SOTA ASR models cannot handle long audios. Feeding them longer recordings, such as 10-minute audio files, would likely degrade transcription accuracy. By breaking the recordings into shorter segments, we aim to align the input format with model training conditions, improving overall transcription quality.

Therefore, we process the data using the following three segmentation approaches among which two are automatic segmentation:  \emph{ Segmented Hybrid Audio Segmentation (SHAS)} \cite{tsiamas2022shasapproachingoptimalsegmentation}, a commonly used segmenter for live transcriptions;  \emph{PyanNet segmentation} \cite{bredin21_interspeech, bain2022whisperx}, a segmentation trained on voice activity detection and further fine-tuned for speaker diarization;  \emph{Human segmentation } serves as a ground truth segmentation.
We provide description on the process of human annotation in Section \ref{sec:manual_seg}. 

 SHAS detects pauses and other acoustic cues to identify natural breakpoints in speech, ensuring that segments correspond to meaningful units of conversation. This approach helps preserve the conversational structure while generating smaller, more manageable audio segments for further processing. 

 In order to achieve a segmentation more aligned with our ground truth as alternative, we use PyanNet segmentation \cite{bredin21_interspeech}.
 To further improve alignment to our scenario, we use a fine-tuned version, of the model, which was trained to track up to three speakers simultaneously in noisy scenarios.
Inspired by WhisperX \cite{bain2022whisperx}, we enforce length constraints through post-processing, where overly long segments are split at their lowest-confidence point, while overly short ones are merged with neighboring segments until the desired segment's duration is achieved.


\section{Evaluation}
We evaluate SOTA ASR systems to establish a  baseline performance on this dataset. 
Through this analysis under varying segmentation and transcription conditions, we identify key challenges that the dataset presents for current ASR technology.

\paragraph*{Metrics}\label{sec:metric}
Word Error Rate (WER) is a common metric used to evaluate the accuracy of ASR systems.
It measures how much the transcribed text deviates from the ground truth by computing the number of errors made during transcription, giving equal importance to every word in the transcript. 
Unlike the other languages in our dataset, Chinese is not a whitespace-separated language. 
We use jieba\footnote{https://github.com/fxsjy/jieba}, a Python Chinese word segmentation tool for segmenting the Chinese text into words.

Our interest also lies in investigating model performances on special words. Additionally,  we observe frequent occurrences of code-switching within the recorded speech and evaluate the model performance. The following provides details of this evaluation:

\begin{itemize}
    \item \textit{Domain-specific WER}: Our dataset comprises scientific conversations in which speakers frequently use domain-specific words.
    We measure the quality of the domain-specific words with respect to the reference and the hypothesis similar to recall and precision \cite{sinhamahapatra2025slideshelpmultimodalcontext}. First, we investigate how many domain-specific words in the reference are missed or wrongly transcribed by the model, by aggregating the deletion and the substitution counts, and dividing it by the total occurrences of domain-specific words in the manual transcript. In this paper, we calculate a reference-centric WER metric $\text{WER}_{t_{ref}}$.
 
 \begin{math}
\mbox{$\text{WER}_{t_{ref}}$}=\frac{|\mbox{substituted} +\mbox{deleted}|} {|\mbox{recognized }+\mbox{substituted}+\mbox{deleted}|}
\end{math}

Next, we calculate the $\text{WER}_{t_{hyp}}$ to evaluate how many domain-specific words in the model's output are incorrectly transcribed.

\begin{math}
\mbox{$\text{WER}_{t_{hyp}}$}=\frac{|\mbox{substituted} +\mbox{inserted}|} {|\mbox{recognized }+\mbox{substituted}+\mbox{inserted}|}
\end{math}
    \item \textit{Code-Switching Performance}: In our multilingual dataset, comprising spontaneous discussions on scientific topics, we observe code-switching behavior of speakers. Non-English speakers often incorporate English words.  One reason for this tendency is the 
absence or limited familiarity of equivalent terminology in their native language. 
To assess code-switching performance, we employ the recently proposed Point-of-Interest Error Rate (PIER) metric \cite{ugan2025pier}. 
PIER is a variant of the traditional Word Error Rate (WER), designed specifically to measure ASR performance on code-switched segments by focusing on errors aligned with points of interest that is, embedded-language words. 
In our case, these annotated English words served as points of interest for PIER computation.
\end{itemize}



\subsection{Evaluation of Different Conditions}
\begin{table*}[h]
  \caption{Overview table with WER on manual (oracle), PyanNet and SHAS 
 segmented audio recordings across the three devices on Whisper}
  \label{tab:overall_tab}
  \centering
  \resizebox{1.0\textwidth}{!}{%
  \begin{tabular}{l|ccc|ccc|ccc}
  
    \toprule
   
         & \multicolumn{3}{c|}{\textbf{Aria}} & \multicolumn{3}{c|}{\textbf{OWL}} & \multicolumn{3}{c}{\textbf{Pi}} \\

     \midrule
    \textbf{segmentation} &  \textbf{manual}   & \textbf{PyanNet}  & \textbf{SHAS}  & \textbf{manual}    & \textbf{PyanNet}   & \textbf{SHAS} & \textbf{manual}   & \textbf{PyanNet} & \textbf{SHAS}\\
  \midrule
    \textbf{ all recordings}& 12.12 & 23.19 & 27.46 & 12.98 & 22.78 & 31.16 & 18.65 & 21.89 & 28.16 \\
    
    \bottomrule
  \end{tabular}}
\end{table*}

As an initial step, we compute the multilingual WER across the entire benchmark under nine distinct recording conditions (Table \ref{tab:overall_tab}). These conditions are derived from the combination of three recording devices and three segmentation strategies: two automatic methods, PyanNet-based diarization and SHAS, and one manual segmentation approach, which serves as the oracle. This experiment is conducted using the SOTA ASR model Whisper.

The results in Table \ref{tab:overall_tab} demonstrates that our benchmark is challenging, with WERs reaching up to 31\%. 
We observe that one of the main challenges is the segmentation of multilingual audio.

When using SHAS segments, WERs range between 27\% and 31\%. In contrast, applying the PyanNet speaker diarization method reduces WERs to approximately 21–23\%. The lowest error rates, between 12\% and approximately 18\%, are achieved by Whisper using our oracle setups. Finally, we observe notable performance variations across the different audio recording conditions, underscoring the impact of recording quality on ASR accuracy.

\subsection{Evaluation of Models for Multilingual Transcription}\label{sec:eval_multi_lingual}
Table~\ref{tab:multi_data_wer} summarizes our findings of this experiment. 
\begin{table}[!h]
  \caption{Evaluation of models for multilingual transcription using manual segments of the OWL recordings. A dash (-) represents scores not considered for the languages, as the corresponding models were not trained on them.}
  \label{tab:multi_data_wer}
  \centering
  \resizebox{1.0\linewidth}{!}{%
  \begin{tabular}{l|c|c|c|c}
  
    \toprule
     \textbf{Language}   &  \textbf{Whisper } & \textbf{SALMONN } & \textbf{Phi }  & \textbf{wav2vec2}
\\
  \midrule

    English         & 10.32  & 17.17 & 16.34  & 31.74 \\ 
    German         & 12.22 & - & 15.72 & 27.93 \\ 
    Turkish         &  15.96 & - &- & 71.24 \\ 
    Chinese         &  14.95 & - & 14.11 & 53.26 \\ 
    Vietnamese         & 24.18  & - &- & 81.84 \\ 
   
    \bottomrule
 \end{tabular}}
\end{table}
The main challenge of the MUSCAT dataset for the models is its  multilingual composition. As a result, we evaluate the ASR performance on our dataset. 
Since it is only possible to separate the languages in the manual segmentation, we use these segments for the experiment. Furthermore, to ensure comparable conditions for both speakers, we focus on segments from the OWL recordings. 

Since SALMONN is trained only on English, the scores for rest of the languages are not considered in this paper.    
The Phi model supports English, German, and Chinese, but does not handle the other languages in our dataset effectively. Among all tested models, Whisper achieves the lowest WER for each language, showing strong multilingual capabilities. 
Table~\ref{tab:multi_data_wer} also demonstrates results for the wav2vec2 model which comparatively has higher WERs than the other models considered for this experiment.

Although  Whisper is performing best on English, we still observe strong performance on all languages with Vietnamese being the most difficult. Similarly, wav2vec2 struggles the most for Vietnamese, Turkish, and Chinese among the five languages. This highlights the importance of having a diverse set of languages in MUSCAT to test the robustness of ASR systems.



\subsection{Evaluation of Recording Devices}\label{sec:eval_recording_device}
In this section we analyse the challenges concerning different recoding quality as  described in Section~\ref{sec:experimental_setup}. For this experiment, we use the SOTA model Whisper and summarize the results in Table~\ref{tab:devices_wer}.
The table shows the WER scores of the transcripts produced by the model across different recording devices. Of three devices, the quality of recordings by Aria may be impacted if the speaker is not wearing it. As a result, we check the model performance separately with and without wearing Aria. To make the distinction explicit, we mark the languages those were recorded with Aria glass in table \ref{tab:devices_wer} with \textit{(Aria)}. For example, the first row of the table shows scores of the recordings when a speaker is not wearing Aria and is speaking English. In contrast, the second row of the table contains scores for recordings where a speaker is speaking English while wearing Aria. 
\begin{table}[ht]
  \caption{Whisper WER across different devices on the manually segmented audio. We mark the languages recorded using Aria glasses in the table with \textit{(Aria)}.}
  \label{tab:devices_wer}
  \centering
  \resizebox{0.8\linewidth}{!}{%
  \begin{tabular}{l|ccc}
  
    \toprule
   
     \textbf{Language}      & \textbf{Aria} & \textbf{OWL}      & \textbf{Pi}       \\
  \midrule
    English         & 9.68  & 8.15 & 12.19   \\
     English \textit{(Aria)}    & 15.06  & 21.21&  39.06 \\
    German \textit{(Aria)}       & 8.71  & 12.22& 14.97   \\
    Turkish         &  16.63 & 15.96 & 23.50   \\
    Chinese \textit{(Aria)}         & 9.26  & 14.95 & 18.74 \\ 
    Vietnamese \textit{(Aria)}         & 26.25  & 24.18& 22.95   \\ 
    \bottomrule
  \end{tabular}}
\end{table}

 The table highlights that the challenges in the different audio conditions vary. If the speaker wears the glass (second row), the Aria microphone clearly leads to the best performance, with quality gains up to 29\% for English (Aria) when compared to the OWL score.
 In contrast, for English (first row) where the Aria is not worn by the speaker, the WER is relatively higher than OWL.  
 This indicates that additional research is needed to also perform good quality ASR without close-source microphones. 
 In addition, Aria microphones perform surprisingly well also for the speaker not wearing the Aria glass (first and fourth row of Table \ref{tab:devices_wer}).
 For these cases we see minor decrease in performance compared to the OWL recordings.

Finally, although both microphones (OWL and Pi) are positioned similarly, there is a clear performance gap. This also motivates additional research on high-quality ASR with lower-quality microphones.


\begin{table*}[ht]
  \caption{An example of ASR performance using two types of automatic segmentation. With the {\color{red}red} color we indicate the substitutions, words omitted from the transcript are shown with {\color{gray}gray}, words in {\color{blue}blue} are inserted texts and, marked with {\color{ForestGreen}green} are the parts where the model is unable to switch to the respective language.}
  \label{tab:example_tab}
  \centering
  \resizebox{1.0\textwidth}{!}{%
  \tiny
  \begin{tabular}{p{3cm}|p{3cm}|p{3cm}}
  
    \toprule
          \textbf{Reference transcript} & \textbf{ASR on SHAS segments}  & \textbf{ASR on PaynNet segments}    \\
    
  \midrule
  \vspace{0.02cm}
  Okay, I have another question. Is this model have the similar architecture as the chatGPT model?
  
  Mehr oder weniger. Es ist ein Transformer, aber es ist so ein bisschen wie bei PaLM, dass die MLP-Schichten und die Attention-Schichten parallel zueinander sind statt sequenziell.
  
  So it's not autoregressive. It's a parallel structure? 
  
  No, no, this is, das ist das ist nur innerhalb von der von einem Transformer-Block. 
  &
  {\transparent{0.4}Okay,} {\color{ForestGreen} Ich habe noch eine Frage. Ist dieses Modell mit der gleichen Architektur wie das HHGPD Modell?} 
  
  Mehr oder weniger. Es ist ein Transformer, aber es ist {\transparent{0.4}so} ein bisschen wie bei {\color{red}Plum}, dass die MLP Schichten und die Attention Schichten parallel zueinander sind, statt {\color{red}sequenziert}. 
  
  So it's not autoregressive, it's a parallel structure? 
  
  No, no, this is{\transparent{0.4}, das ist} {\color{ForestGreen}only inside of one} transformer block. 
  & {\transparent{0.4}Okay,} I have another question. {\color{red}Does} this model have the similar architecture as the chatGPT model?
  
  mehr oder weniger. Es ist ein Transformer, aber es ist so ein bisschen wie bei {\color{red}Plum}, dass die MLP-Schichten und die Attention-Schichten parallel zueinander sind statt sequenziell.
  
  {\transparent{0.4} So it’s not autoregressive.
It’s a parallel structure?} 

{\color{ForestGreen}Nein, nein}, {\color{blue}nein,} {\transparent{0.4} this is,} das ist nur innerhalb von {\transparent{0.4}der von} einem Transformer-Block.  \\

    \bottomrule
  \end{tabular}}
\end{table*}
\subsection{Evaluation of Segmentation Approaches}\label{sec:eval_segmentation_approach}
\begin{table}[ht]
  \caption{Model WER across all languages pairs using OWL  SHAS automatic segments and the manual segments}
  \label{tab:segment_wer}
  \centering
  \resizebox{1.0\linewidth}{!}{%
  \begin{tabular}{l|ccc}
  
    \toprule
     \textbf{}      & \textbf{Whisper} & \textbf{} & \textbf{}            \\
      \midrule
    \textbf{Language pairs}      & \textbf{Manual} & \textbf{PyanNet}& \textbf{SHAS}     \\
   
     \textbf{}      & \textbf{WER} & \textbf{WER}  & \textbf{WER}             \\
  \midrule
    English-German         & 10.88 & 20.57  & 23.93\\ 
    English-Turkish         &  19.89   & 32.53 & 57.41\\ 
    English-Chinese         & 8.16   & 12.89 & 19.29\\ 
    English-Vietnamese        &  12.89  & 24.10 &  31.19\\ 
   
    \bottomrule
  \end{tabular}}
\end{table}

We also investigate the impact of unsegmented audio, which is a more realistic condition, on the final ASR performance. 
Table~\ref{tab:segment_wer} presents results obtained with two automatic segmentation methods and a manual segmentation, which serves as the oracle (described in Section~\ref{subsec:segmentation}). The evaluation is carried out using the Whisper model, which exhibits the strongest multilingual performance across all five dataset languages (Section~\ref{sec:eval_multi_lingual}).

We find that using SHAS segmentation, the model often fails to separate languages properly, resulting in mixed-language segments. This leads to almost three times higher WER for all language pairs, compared to manual segmentation, as the Whisper model struggles with detecting language to transcribe properly within a single segment. For instance,  the SHAS score for English-Turkish recordings is 57.41 which is three times more the manual score 19.89. 

In contrast, using the PyanNet segments which are trained for speaker diarization, leads to fewer segments with speaker overlap. 
As a result, the segments are  more likely to be language homogeneous, aligning better with the pre-trained distribution of single-language utterances.
Consequently, this segmentation approach yields improved ASR performance to SHAS-based automatic segmentation.
The second row of Table~\ref{tab:segment_wer} shows one such example where PyanNet score is lower compared to the SHAS score. 
This highlights the potential of more advanced segmentation techniques to enhance transcription quality in multilingual settings.
The findings emphasize the importance of multilingual datasets such as MUSCAT to benchmark segmentation techniques in ASR.

Table \ref{tab:example_tab} presents an excerpt of a conversation in German and English from MUSCAT, in which two speakers discuss a scientific paper \cite{gunasekar2023textbooks}. 
The first column contains the manually transcribed reference, while the second and third columns show automatic transcriptions generated by Whisper, based on SHAS and PyanNet segmentation, respectively. 
The example demonstrates that the model often fails to detect language switches in the SHAS-based segments, often producing translations instead of transcriptions. 
In contrast, the PyanNet-based segmentation partially mitigates this issue, though it occasionally omits parts of the conversation.


\begin{table}[ht]
  \caption{WER and WER-Term on domain-specific words. Whisper, SALMONN, Phi and wav2vec2 are evaluated using OWL English manual segments.}
  \label{tab:spcial_wer}
  \centering
  \resizebox{1.0\linewidth}{!}{%
  \begin{tabular}{l|c|c|c|c}
    \toprule
  
             &\textbf{Whisper}     & \textbf{SALMONN} & \textbf{Phi} & \textbf{wav2vec2}\\
    \midrule
      Total Counts & 55 & 55 & 55 & 55 \\
     Recognized & 33 & 24 & 19& 4 \\
     Non Recognized & 22 & 31 & 36 & 51 \\
     WER & 10.32 & 17.17 & 16.34 & 31.74\\
     $\text{WER}_{t_{ref}}$  & 35.08 & 46.87 & 59.67 &77.99\\
     $\text{WER}_{t_{hyp}}$ & 28.33 & 46.87 & 59.67 &77.46\\
  %
    \bottomrule
  \end{tabular}}
\end{table}

\subsection{Model Performance on Special Words}\label{sec:eval_special_words}
Our dataset contains multilingual recordings of discussions over scientific papers. Such papers contain technical terms often not found is general discourse, referred to as special words in this paper.  We want to measure the model performance in transcribing such special words, as previous research \cite{wang2024slideavsr} \cite{yang2024mala} has highlighted that such words pose particular challenges for ASR systems. Since the introduced special words in scientific papers are often in English, we focus this experiment only on the English recordings using OWL.

For each recording, we extract domain-specific words from the scientific paper that the speakers use to discuss. To identify these words, we exclude all words found in a general-purpose dataset \cite{di-gangi-etal-2019-must}, from all the words of the paper. The remaining words are considered as the special words associated with the paper.
The \textit{Total Counts} score in Table~\ref{tab:spcial_wer} represents the number of these special words occurring in the English portion of the recordings. The second and third rows of the table indicate the number of instances where the models successfully recognize or failed to recognize these words. 
Finally, we compute the WER on the full vocabulary and the domain-specific WER on the special terms following the evaluation metric outlined in Section~\ref{sec:metric}.

We evaluate the performance of four ASR models Whisper, SALMONN, Phi, and wav2vec2 on domain-specific words. For SALMONN, Phi, and wav2vec2, the $\text{WER}_{t_{ref}}$ and $\text{WER}_{t_{hyp}}$ are approximately 2.3 to 2.7 times higher than the overall WER across all words. In contrast, Whisper exhibits $\text{WER}_{t_{ref}}$, approximately 3.5 times higher than its overall WER, while its $\text{WER}_{t_{hyp}}$ scores are approximately 19\% lower than the corresponding $\text{WER}_{t_{ref}}$ scores. These results illustrate the difficulty current ASR models face when transcribing scientific and technical terms, underscoring the value of the MUSCAT dataset for benchmarking scientific transcription performance.

\begin{table}[ht]
  \caption{ PIER $\downarrow$ on code switched tokens. Whisper, SALMONN, Phi and wav2vec are evaluated using OWL English manual segments.}
  \label{tab:code-switch}
  \centering
  \resizebox{1.0\linewidth}{!}{%
  \begin{tabular}{l|c|c|c|c}
    \toprule
  
          \textbf{Language}   &\textbf{Whisper}     & \textbf{SALMONN} & \textbf{Phi} & \textbf{wav2vec2}\\
    \midrule
     German & 39.29 & 57.14 &64.29  & 116.1 \\
     Turkish & 38.46 & 100.0 &100.0 & 53.85 \\
     Chinese & 77.8  & 66.7 & 77.8 & 88.9 \\
     Vietnamese & 44.76 & 124.76 & 262.86 & 102.91\\

  %
    \bottomrule
  \end{tabular}}
\end{table}

\subsection{Model Performance on Code-Switched Words}
In the MUSCAT dataset, we observe conversational code-switching, where non-English speakers frequently incorporate English words into their speech. Code-switching ASR remains a well-known challenge for current models; therefore, we evaluate the code-switching capabilities of the aforementioned state-of-the-art ASR systems on our dataset using the OWL recording setup.

Our analysis reveals varying degrees of code-switching across languages: Chinese contains the fewest English insertions (9), while Vietnamese exhibits the most (103). Turkish and German speech include 17 and 54 instances, respectively.

As shown in Table~\ref{tab:code-switch}, all models perform significantly worse on code-switched words than on general speech (see Table~\ref{tab:multi_data_wer}). Although the Phi-4 model achieves a WER comparable to Whisper, its PIER score more than doubles, indicating increased difficulty in recognizing embedded English words. The Wav2Vec2 model, which already exhibited the weakest overall performance, remains unsatisfactory under code-switching conditions as well. Overall, we observe that code-switching, while still challenging, appears to be handled slightly better in German speech, possibly due to the linguistic relatedness between German and English. In contrast, Whisper struggles most with Chinese code-switching, whereas multimodally pre-trained models such as SALMONN and Phi-4 perform relatively better on this type of data, likely benefiting from the broader linguistic and acoustic diversity encountered during large-scale pre-training. 

These findings reinforce that code-switching remains a major limitation for current ASR systems and underscore the importance of multilingual, domain-specific datasets such as MUSCAT in advancing research on this challenging phenomenon.

\section{Related Work}
Our work presents a novel dataset that bridges the gap between conversational, multilingual, and academic domains.
Existing general-purpose conversational datasets such as MultiWOZ \cite{budzianowski2018multiwoz}, DialoGPT\cite{DBLP:journals/corr/abs-1911-00536}, \cite{li2017dailydialog} and ConvAI2\cite{dinan2020second}, primarily focus on casual dialogue, including structured interactions or discussions extracted from platforms like Reddit. Other speech datasets include the AMI Meeting Corpus\cite{kraaij2005ami} which consists of meeting recordings and DIPCO\cite{van2019dipco}, a dataset with natural conversation around a dinner table. Both of these speech corpora consist of speech in English. With respect to all these datasets, our work contributes to the field of academic conversational datasets, specifically including one-to-one discussions about scientific papers from known conferences.

Each of the platforms like arXiv, PubMed, and Semantic Scholar primarily contain scientific papers, articles in the form of written text, while our focus is on multilingual speech data. Similarly, multilingual dialogue datasets like FLoRes-101\cite{goyal2022flores}  and CoVoST\cite{wang2020covost}, lack domain-specificity and are mostly text-based. There exist code-switching datasets where multiple languages can be present in the audio such as Arzen\cite{hamed2020arzen}, DECM\cite{ugan-etal-2024-decm}, SEAME\cite{lyu2010seame}. 

Recent advancements in multilingual speech processing have shifted toward spontaneous, multi-party conversational environments. A notable development is the DISPLACE 2024 and 2025 challenge corpora \cite{kalluri2024second}, which provide over 150 hours of multi-speaker, multi-lingual data specifically annotated for speaker and language diarization in overlapping speech scenarios. Furthermore, the MLC-SLM (Multilingual Conversational Speech Language Model) corpus \cite{mu2025summary} introduces a large-scale, 1600-hour benchmark that addresses the complexities of turn-taking and code-switching across 11 languages. These datasets complement established benchmarks like ML-SUPERB 2.0 \cite{shi2024ml}, which expands evaluation to 142 languages to test the cross-lingual generalization of foundational speech models. In addition to the above mentioned challenge-driven datasets, recent large-scale efforts such as SwitchLingua \cite{xie2025switchlingua} have significantly expanded the diversity of conversational corpora. Introduced as a comprehensive multi-ethnic benchmark, SwitchLingua provides over 80 hours of audio from 174 bilingual speakers across 12 languages.
In contrast, our dataset expands the coverage of multilingual conversational data in speech 
where each instance includes dialogues from a discussion on a scientific paper between two speakers speaking separate languages.

\section{Conclusion}
This paper proposes a novel multilingual dataset to evaluate current ASR systems. Our dataset encompasses scientific conversations in five languages, including English, German, Chinese, Turkish, and Vietnamese. Each conversation consists of a paired speech in two languages, one of which is always English, while the other is one of the four remaining languages.

We perform detailed evaluations on several key aspects of speech recognition using various ASR models, including analysis on different recording devices, evaluating  across languages, verifying model capabilities on segmentation, and investigating model performances on domain-specific and code-switched words.


Experimental results from the MUSCAT dataset show that current SOTA ASR systems still face major challenges in handling natural, multilingual scientific discussions. Specifically, our evaluation indicates these models difficulty in accurately detecting when a speaker switches languages during a conversation. When these switches happen, the systems often make mistakes by either translating the speech into the previous language used or completely failing to transcribe the audio segments. Furthermore, the results show that current technology is not yet robust enough to handle the combination of specialized scientific vocabulary, code-switched words or phrases, and different audio conditions found in real-world expert dialogues.

\section{Limitation}
While the MUSCAT dataset provides a novel benchmark for evaluating multilingual scientific conversations, several limitations must be acknowledged. First, the overall scale of the corpus is relatively small, comprising approximately 65 minutes of audio and 9,066 words.
Second, although the dataset encompasses five distinct languages, there is an imbalance in language distribution, with English dominating the conversations. Every recorded interaction involves at least one English speaker, resulting in a word count that is significantly skewed toward English. This asymmetry naturally stems from the domain of the dataset, since the dialogues are entirely based on scientific papers, the speakers often find it more comfortable and precise to articulate complex technical concepts in English.
Finally, our baseline evaluation was inherently constrained by the language capabilities of certain state-of-the-art models; for instance, SALMONN is trained only on English , and Phi-4-multimodal primarily supports English, German, and Chinese.

\section{Acknowledgement}
This work was supported by the European Union’s Horizon Europe Framework Programme under grant agreement No. 101213369, project DVPS (Diversibus Viis Plurima Solvo).

Additional support was provided by KiKIT (Pilot Program for Core-Informatics at KIT) of the Helmholtz Association.

We also acknowledge the use of the HoreKa supercomputer, funded by the Ministry of Science, Research, and the Arts of Baden-Württemberg, and by the Federal Ministry of Education and Research.

\nocite{*}
\section{Bibliographical References}\label{sec:reference}

\bibliographystyle{lrec2026-natbib}
\bibliography{references}


\end{document}